\documentclass[11pt]{article}
\usepackage[margin=1in]{geometry}
\usepackage{times}
\usepackage{amsmath,amssymb}
\usepackage{pgfplots,tikz}
\pgfplotsset{compat=1.17}
\usepackage{graphicx}
\usepackage{subcaption}
\usepackage{hyperref}
\title{JoFormer (Journey-based Transformer): Theory and Empirical Analysis on
the Tiny Shakespeare Dataset\thanks{Certain aspects of this work are the subject
 of a pending patent application.}}
\author{Mahesh Godavarti}
\date{}
\begin{document}
\maketitle

\begin{abstract}
Transformers have demonstrated remarkable success in sequence modeling, yet 
effectively incorporating positional information remains a challenging and 
active area of research. In this paper, we introduce \textbf{JoFormer}, a 
journey-based Transformer architecture grounded in a recently proposed non-
commutative algebra for composing transformations across positions. JoFormer 
represents relative positions through learnable directional transforms that are 
sequentially composed along the input, thereby extending and generalizing 
existing approaches based on relative position representations. We derive the 
JoFormer attention mechanism from first principles and show that it subsumes 
standard methods such as rotary transformations as special cases. To evaluate 
its effectiveness, we compare JoFormer to the RoFormer baseline on the Tiny 
Shakespeare character-level language modeling task. Our results demonstrate that
 JoFormer consistently achieves lower perplexity and faster convergence, 
highlighting the advantages of its more expressive, journey-based treatment of 
position. Notably, the per-token JoFormer is still a primitive, conceptual 
variant with layer-independent angles, yet it already demonstrates strong 
performance—underscoring its promise as a proof of concept for more expressive 
architectures. We conclude by discussing how JoFormer offers a principled 
approach to integrating positional structure into Transformer architectures. The
 code used in this work is available at 
\url{https://github.com/mahesh-godavarti/joformer}.
\end{abstract}

\section{Introduction}
Transformers \cite{vaswani2017} have become the dominant architecture for 
natural language processing and other sequence modeling tasks. A key component 
of Transformer models is the mechanism for encoding token positions, since the 
self-attention operation itself is permutation-invariant. Early Transformers 
used fixed sinusoidal \emph{absolute} positional embeddings \cite{vaswani2017}, 
but subsequent research has found that \emph{relative} position encodings often 
yield better performance \cite{shaw2018}. Various schemes for relative 
positional encoding have been proposed, including adding learned pairwise bias 
terms \cite{shaw2018} or modifying the attention computation to depend on 
relative distances \cite{shaw2018,dai2019}. One recent approach is the Rotary 
Position Embedding (RoPE) introduced by Su et al. \cite{su2021}, which encodes 
absolute positions with a rotation matrix and thereby incorporates relative 
position information into the self-attention formulation. RoPE (and the 
associated RoFormer model) has demonstrated strong results on language tasks by 
enabling Transformers to generalize to longer sequences and to smoothly decay 
attention with distance \cite{su2021}. However, existing relative positional 
encoding methods do not explicitly enforce an underlying algebraic structure for
 composing positions. This can limit the model's ability to capture complex 
structural relationships in sequences.

In this work, we build upon a recent theoretical framework for directional, 
\emph{non-commutative} composition of positional transformations 
\cite{godavarti2025} to develop a new Transformer variant called 
\textbf{JoFormer} (Journey-based Transformer). The key idea is to associate each
 position in the sequence with a learnable linear transform, and to define 
relative positional shifts as the \emph{compositional product} of these 
transforms along the path (or ``journey'') between two positions. We incorporate
 this concept into the Transformer self-attention mechanism: instead of using 
fixed positional encodings or a predefined function of relative distance, 
JoFormer uses a \emph{journey-based operator} $T_{p,q}$ that encodes the 
transformation from position $q$ to position $p$ and modulates the interaction 
between the query at $p$ and the key/value at $q$. This approach provides a 
principled way to inject rich, context-dependent positional biases, while still 
basing attention purely on relative positions (not absolute positions).

We present a formal description of the JoFormer self-attention mechanism and 
show that it subsumes standard relative positional encoding schemes as a special
 case. We then conduct an empirical study on the character-level Tiny 
Shakespeare dataset, comparing JoFormer with a RoFormer baseline model. Our 
experiments demonstrate that JoFormer achieves improved modeling performance 
(lower perplexity) and faster convergence on this small dataset, indicating the 
effectiveness of the journey-based positional encoding even in low-resource 
scenarios. In the following, Section~2 provides the theoretical background for 
JoFormer, Section~3 describes the model architecture and rotational encoding 
scheme, Section~4 details the experimental setup, Section~5 presents results and
 analysis, Section~6 discusses the implications of these results and the JoFormer approach, Section~7 offers a broader perspective on JoFormer's role in sequence modeling, and Section~8 concludes with future directions.

\section{Theoretical Background}
\label{sec:theory}
We first formalize the journey-based positional encoding mechanism that 
underpins JoFormer, following the framework of Godavarti \cite{godavarti2025} 
(Section 6.5). Consider a sequence of length $T$. Each position $i$ is 
represented by a pair $(\mathbf{v}_i,\;R_i)$, where $\mathbf{v}_i \in 
\mathbb{R}^d$ is the content (embedding) vector at position $i$ and $R_i$ is a 
learnable positional transform (a $d \times d$ matrix). For any two positions 
$p$ and $q$, with $q < p$, define the \textbf{relative transform} $T_{p,q}$ as 
the composition of the transforms from $q$ up to $p-1$:
\begin{equation}
\label{eq:T}
    T_{p,q} \;=\; R_q \; R_{q+1} \; \cdots \; R_{p-1}, 
    \qquad 
    T_{q,p} \;=\; T_{p,q}^{-1}, 
    \qquad 
    T_{p,p} \;=\; I~,
\end{equation}
where $I$ is the identity transform. Intuitively, $T_{p,q}$ represents the 
composite transformation for the ``journey'' from position $q$ to position $p$. 
We incorporate this operator into the Transformer's self-attention as follows. 
Let $\mathbf{Q}_p$, $\mathbf{K}_q$, and $\mathbf{V}_q$ denote the query, key, 
and value vectors (for a given attention head) at positions $p$ and $q$ 
respectively. Before computing attention, we \emph{modify the key and value} at 
$q$ as perceived by $p$ using the journey transform:
\begin{equation}
\label{eq:KVtransform}
    \tilde{\mathbf{K}}_{p,q} \;=\; T_{p,q}\,\mathbf{K}_q~, \qquad 
    \tilde{\mathbf{V}}_{p,q} \;=\; T_{p,q}\,\mathbf{V}_q~,
\end{equation}
so that $\tilde{\mathbf{K}}_{p,q}$ and $\tilde{\mathbf{V}}_{p,q}$ are the key 
and value vectors from position $q$, transformed into the ``reference frame’’ of
 position $p$ via the composed positional shift. Self-attention weights are then
 computed using the transformed keys:
\begin{equation}
\label{eq:attn-weight}
    \alpha_{p,q} \;=\; \frac{\exp\!\Big( \frac{\mathbf{Q}_p \cdot 
\tilde{\mathbf{K}}_{p,q}}{\sqrt{d}} \Big)}{\sum_{r=1}^{T} \exp\!\Big( 
\frac{\mathbf{Q}_p \cdot \tilde{\mathbf{K}}_{p,r}}{\sqrt{d}} \Big)}~, 
\end{equation}
and the attention output at $p$ is the weighted sum of transformed values:
\begin{equation}
\label{eq:attn-output}
    \mathbf{O}_p \;=\; \sum_{q=1}^{T} \alpha_{p,q}\,\tilde{\mathbf{V}}_{p,q}~.
\end{equation}
Equations (\ref{eq:attn-weight}) and (\ref{eq:attn-output}) define a self-
attention mechanism in which positional information enters 
\emph{multiplicatively} via $T_{p,q}$, rather than through addition or 
concatenation. Because the transforms $R_i$ need not all commute with each other
 (they are in general non-abelian matrices), this formulation has been described
 as a \textit{non-abelian} self-attention mechanism. Importantly, it reduces to 
standard relative positional attention under certain choices of $R_i$. For 
example, if we take every $R_i$ to be a fixed rotation matrix $R$ (the same for 
all positions), then $T_{p,q} = R\,R\,\cdots R$ ($p-q$ times) $= R^{\,p-q}$, 
which depends only on the distance $(p-q)$. In this case, the above equations 
recover the usual Transformer with relative position encodings (such as 
sinusoidal or rotary embeddings). In general, however, each $R_i$ can be learned
 independently, enabling flexible and context-dependent positional biases while 
still keeping the attention mechanism dependent only on relative positions. This
 generality allows JoFormer to adapt the positional transformations to the data,
 potentially capturing patterns (e.g. syntactic or semantic boundaries) that a 
fixed scheme might miss.

\section{Model Architecture and Rotational Encoding}
All three model variants – RoFormer, fixed-angles JoFormer, and per-token 
JoFormer – share an identical backbone architecture, differing only in their 
embedding and self-attention modules. At a high level, each model is a 
Transformer-based autoregressive language model with $N$ layers. Each layer 
consists of a single-head self-attention sublayer followed by a position-wise 
feed-forward network (FFN), with residual connections and layer normalization 
applied around each sublayer. Input tokens are first mapped to embedding 
vectors, which are then processed through these $N$ transformer blocks. The 
overall architecture (number of layers, hidden dimension $d$, etc.) remains the 
same for all variants; the crucial differences lie in how positional information
 is encoded and integrated via a rotation matrix within the self-attention 
mechanism. 

\textbf{Shared Architecture Overview:} In all variants, the self-attention 
module computes query, key, and value vectors ($\mathbf{q}_i$, $\mathbf{k}_i$, 
$\mathbf{v}_i$) for each token at position $i$. Positional information is 
incorporated by rotating these vectors using a position-dependent rotation 
matrix $R_i$. The matrix $R_i$ is defined to act on the $d$-dimensional 
representation by partitioning it into $d/2$ two-dimensional subspaces (each 
corresponding to a pair of hidden dimensions). In each 2D subspace $j$ (covering
  coordinates $2j$ and $2j+1$ of the vector), $R_i$ performs a planar rotation by
  an angle $\phi_{i,j}$ specific to position $i$ and subspace $j$. We can write 
the rotation on the $j$-th pair of dimensions as:
\begin{equation}
R_i^{(j)} \;=\;
\begin{pmatrix}
\cos(\phi_{i,j}) & -\sin(\phi_{i,j})\\
\sin(\phi_{i,j}) & \cos(\phi_{i,j})
\end{pmatrix},
\end{equation}
and the full rotation matrix is the block-diagonal composition of these 
submatrices: 
\begin{equation}
R_i = \mathrm{diag}\!\big(R_i^{(0)}, R_i^{(1)}, \dots, R_i^{(d/2-1)}\big).
\end{equation} 
This $R_i$ transforms a vector by rotating each pair of coordinates 
$(2j,2j{+}1)$ by angle $\phi_{i,j}$. The strategy for determining $\phi_{i,j}$, 
and how $R_i$ is applied within self-attention, differs for each model variant 
as detailed below. 
\paragraph{RoFormer (fixed sinusoidal rotation).} In the RoFormer baseline, the 
rotation angles $\phi_{i,j} = \phi_j$ are predetermined and independent of 
postion $i$. In other words, $R_i = R$ for all $i$ and $T_{p,q} = R^{q-p}$. The 
difference from the formulation in \ref{sec:theory} lies in how $\mathbf{O}_p$ 
is computed:
\begin{equation}
    \mathbf{O}_p \;=\; \sum_{q=1}^{T} \alpha_{p,q}\,\mathbf{V}_{p,q}~.
\end{equation}
\paragraph{Fixed-angles JoFormer.} The fixed-angles JoFormer variant extends the
 RoFormer approach where the key difference is in how $\mathbf{O}_p$ is 
computed:
\begin{equation}
    \mathbf{O}_p \;=\; \sum_{q=1}^{T} \alpha_{p,q}\,\tilde{\mathbf{V}}_{p,q}~.
\end{equation}
\paragraph{Per-token JoFormer (learned rotations).} The per-token JoFormer 
extends the fixed-angles JoFormer by introducing learnable rotation angles on a 
per-token basis, instead of using a fixed positional function. In this variant, 
each token in the vocabulary is associated with a learned angle vector $\theta_i
 \in \mathbb{R}^{d/2}$ as part of its embedding. When a token $x_i$ appears at 
position $i$, we obtain its corresponding angle vector $\theta_i$ (in addition 
to a standard learnable content embedding for the token). From $\theta_i$, we 
construct a token-specific rotation matrix $R(\theta_i)$ (following the same 
block-diagonal $2\times2$ structure, where each subspace $j$ uses angle 
$\theta_{i,j}$). 

\section{Experimental Setup}
\label{sec:experiments}
We evaluate the proposed JoFormer architecture on a character-level language
modeling task using the \textit{Tiny Shakespeare} dataset. This dataset consists
of approximately 1~million characters of Shakespearean text (extracted from
public-domain plays), and is a standard benchmark for testing small-scale
language models. The task is to predict the next character given a context of
previous characters; model performance is measured by the cross-entropy loss and
the corresponding perplexity on a held-out validation set. 

\paragraph{Model Variants and Baseline.} We evaluate three model variants in our
experiments: the baseline \textbf{RoFormer} \cite{su2021} with rotary position
embeddings, and two versions of \textbf{JoFormer} – a
\textbf{fixed-angles JoFormer} and a \textbf{per-token JoFormer}. All variants
share identical architecture hyperparameters. We conduct experiments with
$N=1$, $3$, and $6$ Transformer layers, using a model dimension $d=90$,
a feed-forward inner dimension of $4 \times d$, and a single attention head. The
maximum context length is set to 20 characters. The baseline RoFormer applies
the standard rotary positional encoding in each self-attention layer as
described by Su et al.\cite{su2021}, rotating each query and key vector by a
fixed position-dependent angle. In the fixed-angles JoFormer, we use the same
sinusoidal rotation angles as in RoFormer, but apply the rotation matrix $R_i$
to all components of self-attention (including the value vectors) via the
journey-based approach (see Section\ref{sec:theory}). This variant thus retains
a fixed (non-learned) positional encoding, integrating it through JoFormer's
composition mechanism. In the per-token JoFormer, the rotation angles
$\phi_{i,j}$ are instead learned for each token rather than being fixed
functions of position – i.e., every token is associated with a learnable
rotation vector as part of its embedding, yielding a token-specific $R_i$ at
each position. Aside from these encoding differences, all other model settings
and initialization remain the same across variants. We implement JoFormer's
rotational encoding with minimal overhead: the inner product $\mathbf{Q}_p \cdot (T_{p,q}
\mathbf{K}_q)$ is computed by pre-rotating $\mathbf{Q}_p$ and $\mathbf{K}_q$ by their respective cumulative
angles (analogous to RoPE), so that the journey-based attention computation has
essentially the same complexity as the baseline. As a result, JoFormer incurs
negligible extra cost beyond the storage of the additional rotation parameters.
The number of extra parameters introduced
by fixed-angles JoFormer is \textbf{zero} and the number of extra parameters 
introduced 
by the per-token JoFormer is on the order of $d \times d$; for example, with 
$d=90$,
this amounts to $\sim 8100$ additional parameters.

\paragraph{Training Details.} We train all model variants from scratch on the
Tiny Shakespeare corpus. The text is first tokenized at the character level
(vocabulary of 65 unique characters including letters, digits, punctuation and
whitespace). We use 90\% of the data for training and 10\% for validation.
Models are trained for 10000 epochs using a batch size of 32 sequences (each of
length 20 characters). We employ the Adam optimizer with an initial learning
rate of $3\times 10^{-4}$, and apply a cosine learning rate decay schedule over
the course of training. To reduce variability, we run each experiment with 3
different random initializations and report the average performance. During
training, we monitor the training loss and validation perplexity every epoch. 

\section{Results and Analysis}
\begin{figure}[t]
\centering
\begin{subfigure}{0.32\textwidth}
\centering
\includegraphics[width=\textwidth]{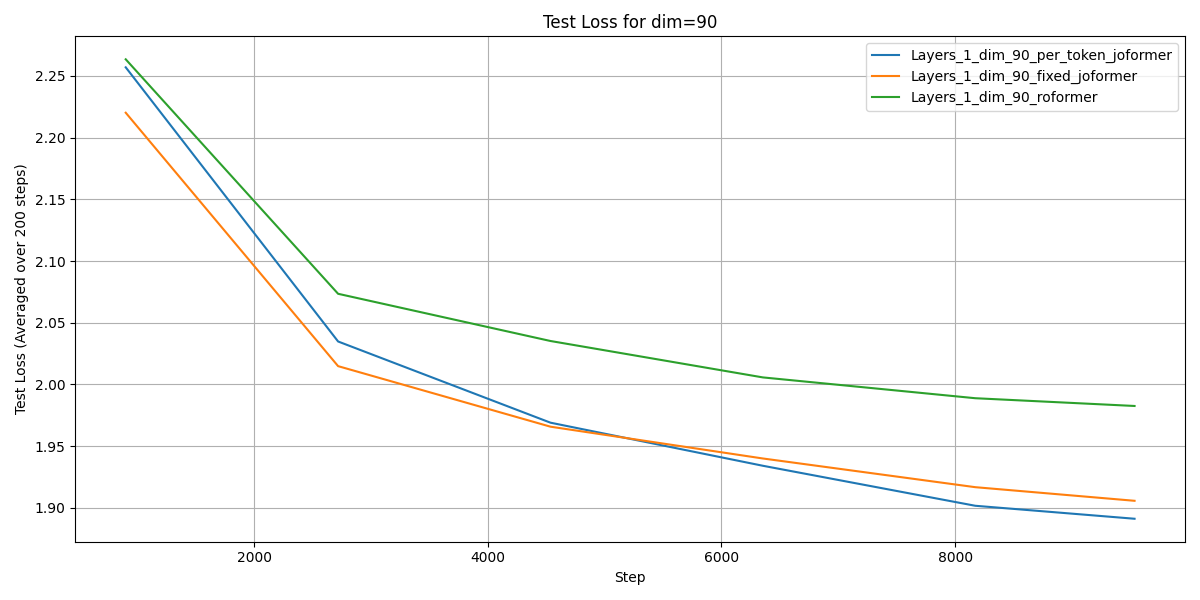}
\caption{1 layer}
\label{fig:val_loss_1layer}
\end{subfigure}
\hfill
\begin{subfigure}{0.32\textwidth}
\centering
\includegraphics[width=\textwidth]{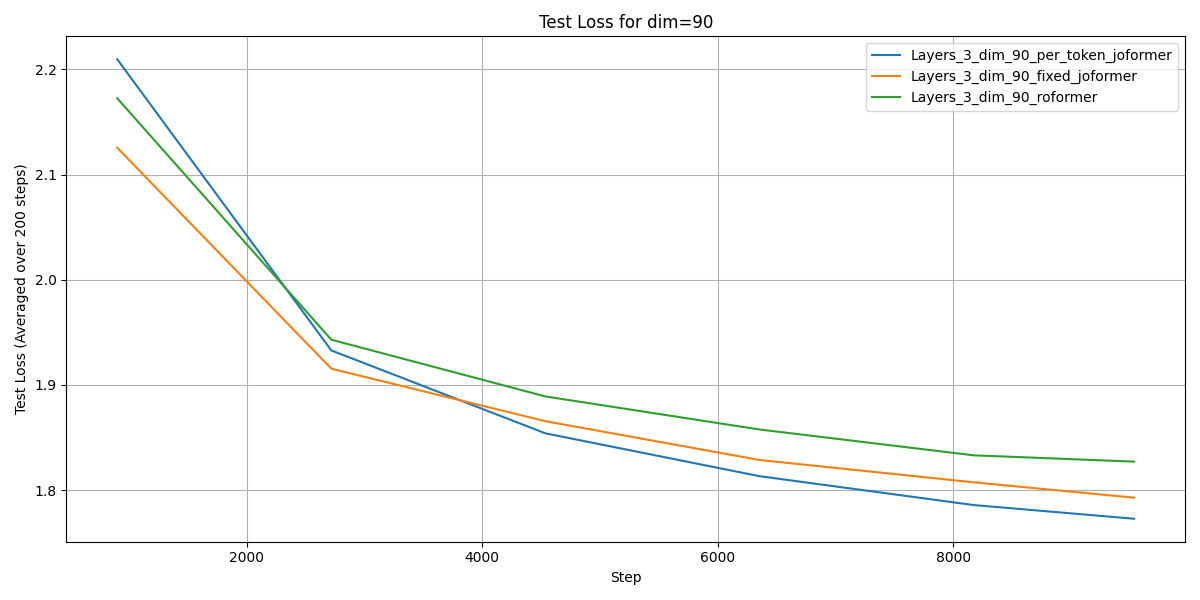}
\caption{3 layers}
\label{fig:val_loss_3layers}
\end{subfigure}
\hfill
\begin{subfigure}{0.32\textwidth}
\centering
\includegraphics[width=\textwidth]{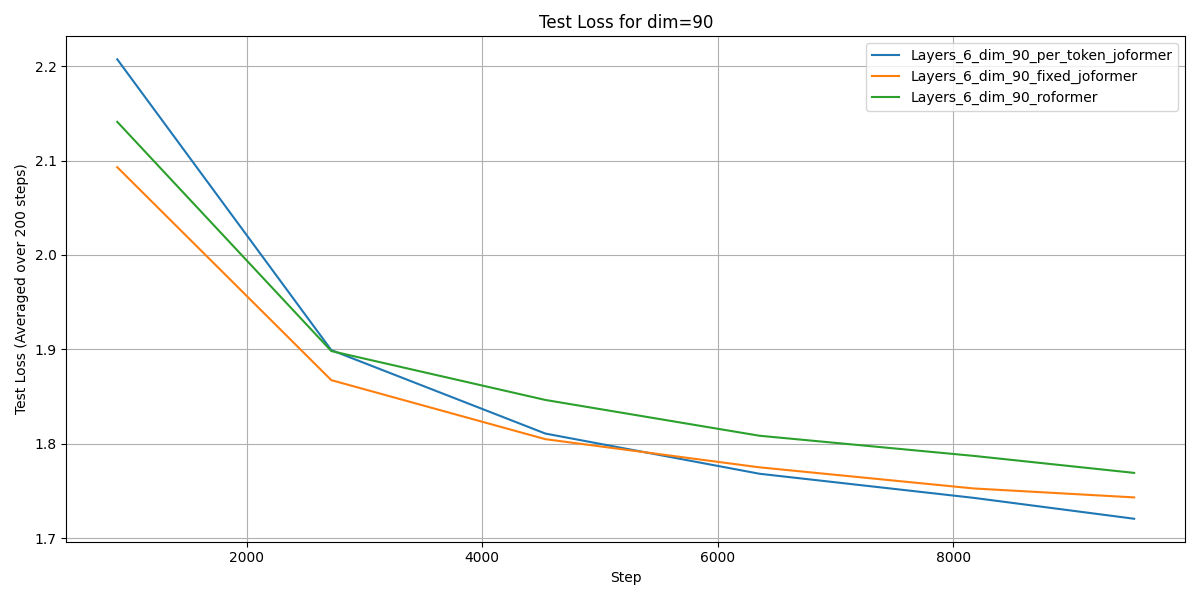}
\caption{6 layers}
\label{fig:val_loss_6layers}
\end{subfigure}
\caption{Validation loss (cross-entropy per token) vs. training epoch for 
RoFormer (baseline) and the JoFormer variants at different model depths. 
\textbf{(a)} With a single layer, the per-token JoFormer exhibits a 
significantly lower validation loss throughout training, outperforming both the 
RoFormer and fixed-angles JoFormer. \textbf{(b)} With three layers, both JoFormer
 variants maintain a lead over RoFormer, though the gap is smaller than in the 
1-layer case; the per-token JoFormer eventually surpasses the fixed-angles 
variant to reach the lowest loss. \textbf{(c)} With six layers, the fixed-angles 
and per-token JoFormer achieve similar performance and continue to outperform 
RoFormer. JoFormer's advantage persists at this depth, but the relative gap is 
reduced compared to shallower models.}
\label{fig:val_loss_depths}
\end{figure}

After training, we observe that both JoFormer variants consistently outperform the RoFormer baseline across all tested model depths. Figure~\ref{fig:val_loss_depths} shows the validation loss (cross-entropy per token) over training epochs for models with 1, 3, and 6 layers. The most significant performance gap appears in the single-layer setting (Figure~\ref{fig:val_loss_1layer}), where the per-token JoFormer ultimately achieves the lowest validation loss despite starting with a higher one—demonstrating the strength of its learnable, token-specific rotations. The fixed-angles JoFormer also improves upon the RoFormer in this shallow configuration, though it trails slightly behind the per-token variant.

As model depth increases to three layers (Figure~\ref{fig:val_loss_3layers}), the performance differences between models narrow, but JoFormer retains its edge. The fixed-angles variant exhibits faster convergence and maintains lower loss than RoFormer throughout training. While the per-token JoFormer again begins with a higher loss, it eventually surpasses both alternatives by the end of training.

At six layers (Figure~\ref{fig:val_loss_6layers}), all three models converge to similar final validation losses, but JoFormer variants still show a small yet consistent advantage in terms of final validation loss. The fixed-angles variant maintains a constant advantage over RoFormer throughout training. This trend confirms that JoFormer's benefits are most pronounced in shallow networks, where its inductive bias provides substantial guidance, while deeper models gradually compensate through brute-force learning. Table~\ref{tab:results_summary} summarizes the final validation perplexities, confirming JoFormer’s consistent improvements over the baseline at every depth.

 \begin{table}[t]
\centering
\caption{Final validation perplexity after 10000 training epochs (lower is 
better), averaged over 3 runs, for each model variant at different depths.}
\begin{tabular}{lccc}
\hline
\textbf{Model} & \textbf{1 Layer} & \textbf{3 Layers} & \textbf{6 Layers} \\
\hline
RoFormer (baseline) & 3.94 & 3.56 & 3.38 \\
Fixed-angle JoFormer & 3.73 & 3.46 & 3.34 \\
Per-token JoFormer & 3.7 & 3.41 & 3.29 \\
\hline
\end{tabular}
\label{tab:results_summary}
\end{table} These results suggest that JoFormer's improvements stem from its 
more expressive
positional encoding mechanism. In particular, the per-token JoFormer—by
learning token-specific positional rotations—can adapt to the data and capture
fine-grained positional biases that a fixed encoding cannot. Anecdotally, we
observed that the per-token JoFormer sometimes places higher attention weights
on distant but syntactically related characters (such as matching quotation
marks or parentheses) compared to the RoFormer. This suggests that the learned
$R_i$ transformations enable the model to encode meaningful long-range “jumps”
or dependencies that are missed by a fixed sinusoidal scheme. Crucially, the 
added flexibility of JoFormer did not lead to overfitting on this
relatively small dataset. All models converged to similar training losses, yet
the JoFormer variants maintained a gap in validation performance, indicating
better generalization. We also note that the computational complexity of
JoFormer's attention mechanism is essentially the same as in RoFormer: the
$T_{p,q}$ rotations are applied implicitly via element-wise operations (rotating
the query and key vectors), so inference speed and memory usage remain virtually
unchanged aside from the negligible increase in parameters. Overall, these 
results validate the effectiveness of the journey-based attention
mechanism in practice. Even on the Tiny Shakespeare dataset, JoFormer's enhanced
positional encoding yields faster convergence and a lower final loss compared to
the standard rotary embedding baseline. This demonstrates that even small-scale
Transformers can benefit from a theoretically grounded, more expressive
positional encoding scheme. \section{Discussion}
Although the absolute perplexity reductions achieved by JoFormer may appear modest, they are quite notable given the Tiny Shakespeare dataset’s low baseline perplexity. For instance, in the 6-layer setting the RoFormer baseline reaches a validation perplexity of 3.38, whereas the per-token JoFormer attains 3.29 (Table~\ref{tab:results_summary}). Reducing perplexity further in such a low-perplexity regime is challenging, so these improvements underscore the efficacy of the journey-based positional encoding even in this small-scale scenario.

Another notable finding is the diminishing performance gap between JoFormer and the baseline as the number of layers increases. In shallower models (e.g., 1 layer), JoFormer provides a large gain over the RoFormer, but with deeper models (6 layers) the gap narrows (though it never vanishes). This trend suggests that adding more layers in a standard Transformer can eventually compensate for the absence of JoFormer's structured positional bias – essentially a brute-force way of learning positional structure through increased depth. By contrast, JoFormer identifies and integrates this structural information with far fewer layers by explicitly encoding the sequence's positional “journey.” In effect, what many layers of a vanilla model might learn implicitly, JoFormer captures through its rotation mechanism in a more direct and data-efficient manner.

Finally, we clarify that the fully general per-token JoFormer remains conceptual at this stage. In our current implementation, the rotation angles $\phi_{i,j}$ are independent of the layer – the same token-specific rotation $R_i$ is applied at every layer of the Transformer. Defining how these positional rotations should vary with layer depth is an open question left for future work. A true per-token, per-layer encoding scheme could further enhance JoFormer's expressiveness, but realizing it will require careful design to determine how angle parameters should evolve across layers.

\section{Bridging Structured State Space Models and Transformers}

In the broader context of sequence modeling, JoFormer bridges the gap between classical Structured State Space Models (SSMs) and Transformers. SSMs (such as S4 or Mamba) maintain a sequentially evolving latent state that provides a strong inductive bias for temporal dependencies, often excelling on tasks requiring explicit long-range sequential dynamics. However, this recurrent state evolution means new inputs are continually integrated into a fixed-size latent state, creating a potential information bottleneck. While this approach is memory-efficient and enforces a smooth sequential structure, it limits the model’s ability to retain and access detailed information from arbitrary earlier positions. As a result, SSMs—despite their strengths on certain long-range tasks (e.g. Long Range Arena benchmarks)—often underperform attention-based Transformers on tasks like language modeling that require integrating rich, global context. Transformers, by contrast, avoid the single-state bottleneck entirely. The self-attention mechanism allows each token to attend directly to any other token in the sequence, enabling dynamic, content-based interactions without relying on a recurrent hidden state. This freedom to focus on relevant tokens gives Transformers greater expressive power: important features or dependencies can be captured via direct token-to-token attention instead of being compressed into a single state. Moreover, self-attention is highly parallelizable across positions, allowing Transformers to scale efficiently on modern hardware and achieve state-of-the-art performance across many domains. The trade-off is that Transformers lack the inherent sequential inductive bias of recurrence and incur a higher quadratic cost in sequence length, but in practice their flexibility to model global context often outweighs these drawbacks for complex tasks. JoFormer’s hybrid approach aims to capture the best of both worlds. It introduces learned per-token rotation matrices that evolve each token’s representation based on its position, analogous to updating a hidden state at every step (imparting an SSM-like sequential bias). As the model processes the sequence, each token embedding is rotated by a learned transformation dependent on its position in the sequence. Crucially, after applying these positional rotations, JoFormer allows information to flow freely between tokens via standard self-attention. In effect, the model is gently biased to account for sequential progression (through the cumulative rotations) while avoiding the restrictive gating of information through a single state vector. All token rotations can be computed in parallel, and their sequential composition (the product of rotations from the start of the sequence up to a given token) does not impede parallel computation. Thus, JoFormer injects a notion of sequential structure into the model’s representations yet preserves the Transformer's flexibility in content-based attention, avoiding the representational bottleneck of classical SSMs. Mathematical Perspective on Weighting and Propagation: Beyond the conceptual view, we can contrast SSMs, Transformers, and JoFormer by examining how each architecture propagates and weights information across positions. In an SSM, the output at position $k$ is given by:
\begin{equation}
y_k = \sum_{i \leq k} C_k \left( \prod_{j=i}^{k-1} A_j \right) B_i x_i
\end{equation}
Here each input $x_i$ contributes to the output $y_k$ scaled by a fixed product of matrices $B_i$, $A_j$ (for $i \le j < k$), and $C_k$ determined by the model’s parameters. These coefficients are content-agnostic – they depend only on position and are fixed in advance, not on the data. This means every token’s influence on future outputs is predetermined: even inputs that turn out to be irrelevant are still propagated with some weight. Such fixed, uniform mixing of information provides a strong inductive bias for smooth sequential processing, but it lacks flexibility in filtering out unimportant information. In other words, an SSM cannot selectively amplify or ignore individual token contributions based on content, which can lead to an information bottleneck as the sequence’s details are compressed into a small latent state. By contrast, a vanilla Transformer computes the output $y_k$ as a content-weighted sum of values:
\begin{equation}
y_k = \sum_{i \leq k} \alpha_{ik} V_i
\end{equation}
Here $\alpha_{ik}$ is the attention weight assigned to value $V_i$ for producing the output at position $k$. These attention weights $\alpha_{ik}$ are data-dependent, computed from dot-products between queries and keys and normalized (e.g. via softmax). Consequently, the model can emphasize important tokens (assigning them large $\alpha_{ik}$ values) and down-weight or ignore irrelevant tokens (assigning near-zero weights). Each output position thus selectively focuses on the most relevant inputs instead of uniformly mixing information from all past tokens. In other words, Transformers bypass the SSM’s fixed-state bottleneck: all token representations are maintained in parallel, and the model dynamically decides which information to use at each step. The flip side is that Transformers have no built-in notion of a recurrent state (aside from explicit positional encodings), but this freedom allows them to learn long-range dependencies purely from data, leveraging content to guide information flow. JoFormer combines these two mechanisms. Formally, JoFormer’s output can be expressed as:
\begin{equation}
y_k = \sum_{i \leq k} \alpha_{ik} \left( \prod_{j=i}^{k-1} R_j \right) V_i
\end{equation}
In this formulation, $R_j$ are learned position-wise transformation matrices applied between positions $i$ and $k$. Like a Transformer, JoFormer uses attention weights $\alpha_{ik}$ to selectively weight each value $V_i$ based on the sequence’s content. However, similar to an SSM, the contribution of $V_i$ is further modulated by the sequential product $\prod_{j=i}^{k-1} R_j$, which represents a learned positional transform applied as the value is carried forward from position $i$ to $k$. Intuitively, one can think of this as the value being gradually transformed (e.g. rotated, decayed, or otherwise evolved) as it travels through intermediate positions, much like an SSM’s state update $A_j$ would propagate information through time. Importantly, because the attention coefficient $\alpha_{ik}$ can be zero for many $i$, JoFormer is not forced to propagate every token’s information through the entire chain of $R_j$ transformations. It only carries forward the contributions of tokens deemed relevant by attention, thereby avoiding the indiscriminate accumulation of all tokens’ information. In effect, JoFormer preserves the salient information pathways (via content-based attention) while still applying a structured, position-dependent transformation along those paths (via the $R_j$ matrices). This approach aligns with the intuition of combining the efficiency and inductive bias of recurrent architectures with the adaptive, content-aware processing of attention-based models. By construction, JoFormer’s design mitigates the limitations of both SSMs and Transformers. It circumvents the SSM’s fixed-weight bottleneck by allowing data-dependent gating of information through attention, and it instills into the Transformer’s free-form attention an SSM-like sequential prior that guides learning. 

In essence, JoFormer marries the strengths of the two paradigms: it retains the Transformer's ability to focus on important content and integrate global context, while imparting an explicit sequential structure akin to an SSM. This balanced hybrid of filtering and sequential propagation yields a more expressive sequence modeling capability, offering both long-range modeling efficiency and flexible context integration.

\section{Conclusion}
In this paper, we presented JoFormer, a journey-based Transformer architecture
that integrates a non-commutative positional encoding mechanism into the self-
attention framework. The core idea was to associate each position with a
learnable transform and compose these transforms to obtain relative “journey”
operators between positions. We derived this formulation from a theoretical
algebraic framework and showed that it generalizes existing relative positional
encoding methods (recovering rotary embeddings as a special case). Empirically,
our experiments on a character-level language modeling task (Tiny Shakespeare)
demonstrated that JoFormer can achieve lower perplexity and faster convergence
than a strong Transformer baseline with rotary positional embeddings. 

This work opens up several directions for future research. First, it will be
important to validate JoFormer’s performance on larger-scale datasets and tasks
(such as word-level language modeling or machine translation) to confirm that
the benefits hold in more complex settings. Second, the journey-based approach
naturally extends to data with multiple positional axes or dimensions (e.g. 2D
grids for images, or tree-structured data). Exploring JoFormer variants for
image modeling or other modalities could unveil further advantages of the
proposed framework. Finally, analyzing the learned positional transforms $R_i$
could provide insights into how the model encodes positional information — for
example, whether certain transforms correspond to known linguistic structures or
repetitive patterns in text. We believe that incorporating structured
positional priors via the journey-based mechanism is a promising avenue for
enhancing Transformer models, and we hope this work inspires further
developments in this direction.  
\end{document}